\documentclass[11pt,a4paper]{article}
\usepackage[latin9]{inputenc}
\usepackage{amsmath}
\usepackage{amssymb}
\usepackage{graphicx}
\PassOptionsToPackage{normalem}{ulem}
\usepackage{ulem}

\makeatletter

\pdfpageheight\paperheight
\pdfpagewidth\paperwidth

\providecommand{\tabularnewline}{\\}

\@ifundefined{date}{}{\date{}}
\usepackage[hyperref]{acl2019}
\usepackage{times}
\usepackage{latexsym}

\usepackage{url}
\usepackage{algorithm}
\usepackage{algpseudocode}
\usepackage{multirow}

\aclfinalcopy % Uncomment this line for the final submission
\title{Automated Chess Commentator Powered by Neural Chess Engine}

\author{Hongyu Zang\thanks{\enskip The two authors contributed  equally to this paper.} \and Zhiwei Yu\footnotemark[1] \and Xiaojun Wan\\
        Institute of Computer Science and Technology, Peking University \\
        The MOE Key Laboratory of Computational Linguistics, Peking University \\ 
        Center for Data Science, Peking University\\ \texttt{\{zanghy, yuzw, wanxiaojun\}@pku.edu.cn}}

\makeatother

\begin{document}
\maketitle 
\begin{abstract}
In this paper, we explore a new approach for automated chess commentary
generation, which aims to generate chess commentary texts in different
categories (e.g., \textit{description}, \textit{comparison}, \textit{planning},
etc.). We introduce a neural chess engine into text generation models
to help with encoding boards, predicting moves, and analyzing situations.
By jointly training the neural chess engine and the generation models
for different categories, the models become more effective. We conduct
experiments on 5 categories in a benchmark Chess Commentary dataset
and achieve inspiring results in both automatic and human evaluations. 
\end{abstract}

\section{Introduction}

\begin{figure}
\begin{centering}
\includegraphics[width=3in]{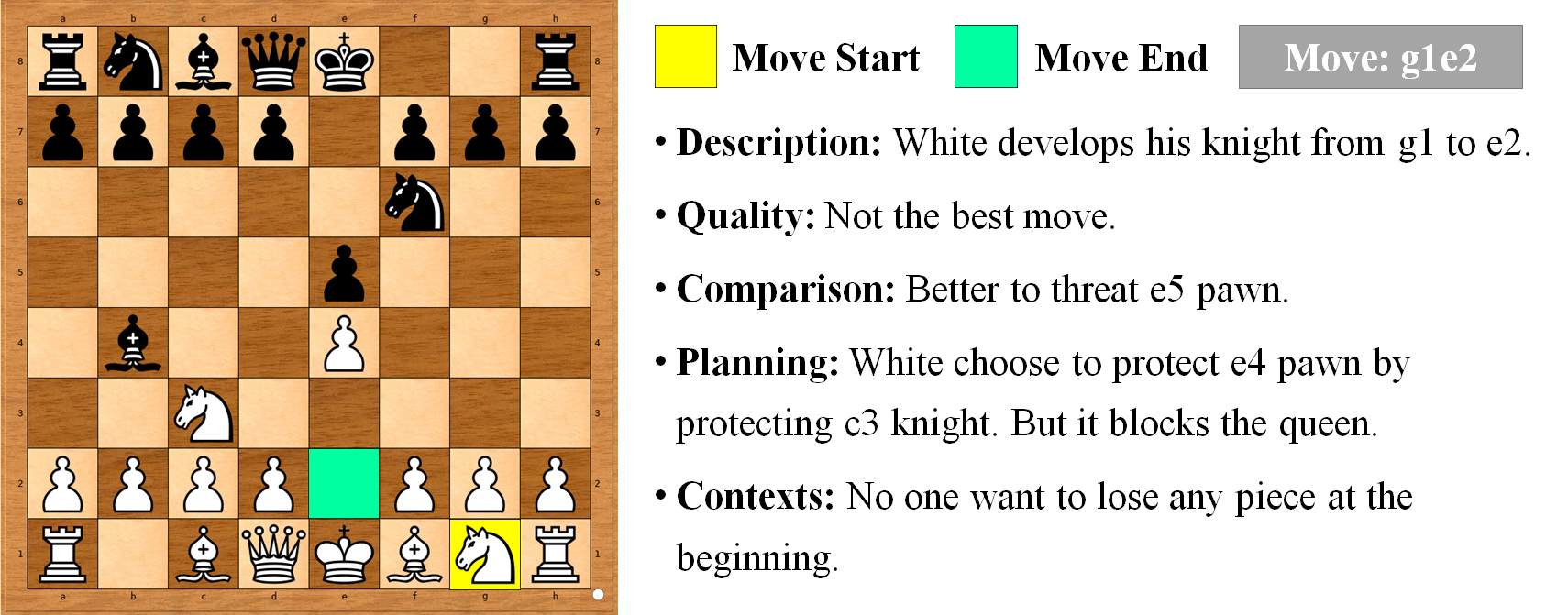} 
\par\end{centering}
\caption{Chess Commentary Examples. \label{fig:intro_board}}
\end{figure}

With games exploding in popularity, the demand for Natural Language
Generation (NLG) applications for games is growing rapidly. Related
researches about generating real-time game reports \cite{DBLP:conf/inlg/YaoZWX17},
comments \cite{DBLP:conf/acl/HovyNBGJ18,DBLP:conf/cig/KamekoMT15},
and tutorials \cite{DBLP:conf/fdg/GreenKBNT18,DBLP:journals/corr/abs-1805-11768}
benefit people with entertainments and learning materials. Among these,
chess commentary is a typical task. As illustrated in Figure \ref{fig:intro_board},
the commentators need to understand the current board and move. And
then they comment about the current move (\textit{Description}), their
judgment about the move (\textit{Quality}), the game situation for
both sides (\textit{Contexts}), their analysis (\textit{Comparison})
and guesses about player's strategy (\textit{Planning}). The comments
provide valuable information about what is going on and what will
happen. Such information not only make the game more enjoyable for
the viewers, but also help them learn to think and play. Our task
is to design automated generation model to address all the 5 sub-tasks
(\textit{Description}, \textit{Quality}, \textit{Comparison}, \textit{Planning},
and \textit{Contexts}) of single-move chess commentary.

Automatically generating chess comments draws attention from researchers
for a long time. Traditional template-based methods \cite{ACT} are
precise but limited in template variety. With the development of deep
learning, data-driven methods using neural networks are proposed to
produce comments with high quality and flexibility. However, generating
insightful comments (e.g., to explain why a move is better than the
others) is still very challenging. Current neural approaches \cite{DBLP:conf/cig/KamekoMT15,DBLP:conf/acl/HovyNBGJ18}
get semantic representations from raw boards, moves, and evaluation
information (threats and scores) from external chess engines. Such
methods can easily ground comments to current boards and moves. But
they cannot provide sufficient analysis on what will happen next in
the game. Although external features are provided by powerful chess
engines, the features are not in a continuous space, which may be
not very suitable for context modeling and commentary generation.

It is common knowledge that professional game commentators are usually
game players. And expert players can usually provide more thorough
analysis than amateurs. Inspired by this, we argue that for chess
commentary generation, the generation model needs to know how to think
and play in order to provide better outputs. In this paper, we introduce
a neural chess engine into our generation models. The chess engine
is pre-trained by supervised expert games collected from FICS Database\footnote{https://www.ficsgames.org/\label{fics}}
and unsupervised self-play \cite{DBLP:journals/corr/abs-1712-01815,silver2017mastering}
games, and then jointly trained with the generation models. It is
able to get board representations, predict reasonable move distributions,
and give continuous predictions by self-play. Our generation models
are designed to imitate commentators' thinking process by using the
representations and predictions from the internal chess engine. And
then the models ground commentary texts to the thinking results (semantics).
We perform our experiments on 5 categories (\textit{Description},
\textit{Quality}, \textit{Contexts}, \textit{Comparison}, \textit{Planning})
in the benchmark Chess Commentary dataset provided by Jhamtani \shortcite{DBLP:conf/acl/HovyNBGJ18}.
We tried models with different chess engines having different playing
strength. Both automatic and human evaluation results show the efficacy
and superiority of our proposed models.

The contributions are summarized as follows: 
\begin{itemize}
\item To the best of our knowledge, we are the first to introduce a compatible
neural chess engine to the chess comment generation models and jointly
train them, which enables the generation models benefit a lot from
internal representations of game playing and analysis. 
\item On all the 5 categories in the Chess Commentary dataset, our proposed
model performs significantly better than previous state-of-the-art
models. 
\item Our codes for models and data processing will be released on GitHub\footnote{https://github.com/zhyack/SCC}.
Experiments can be easily reproduced and extended. 
\end{itemize}

\section{Related Works}

The most relevant work is \cite{DBLP:conf/acl/HovyNBGJ18}. The authors
released the Chess Commentary dataset with the state-of-the-art Game
Aware Commentary (GAC) generation models. Their models generate comments
with extracted features from powerful search-based chess engines.
We follow their work to further explore better solutions on different
sub-tasks (categories) in their dataset. Another relevant research
about Shogi (a similar board game to chess) commentary generation
is from Kameko et al. \shortcite{DBLP:conf/cig/KamekoMT15}. They
rely on external tools to extract key words first, and then generate
comments with respect to the key words. Different from their works,
in this paper, we argue that an internal neural chess engine can provide
better information about the game states, options and developments.
And we design reasonable models and sufficient experiments to support
our proposal.

Chess engine has been researched for decades \cite{levy1982computers,DBLP:journals/ml/BaxterTW00,DBLP:journals/corr/abs-1711-09667,DBLP:journals/corr/abs-1712-01815}.
Powerful chess engines have already achieved much better game strength
than human-beings \cite{campbell2002deep,DBLP:journals/corr/abs-1712-01815}.
Traditional chess engines are based on rules and heuristic searches
\cite{marsland1987computer,campbell2002deep}. They are powerful,
but limited to the human-designed value functions. In recent years,
neural models \cite{silver2016mastering,silver2017mastering,DBLP:journals/corr/abs-1711-09667}
show their unlimited potential in board games. Several models are
proposed and can easily beat the best human players in Go, Chess,
Shogi, etc. \cite{DBLP:journals/corr/abs-1712-01815}. Compared to
the traditional engines, the hidden states of neural engines can provide
vast information about the game and have the potential to be compatible
in NLG models. We follow the advanced techniques and design our neural
chess engine. Apart from learning to play the game, our engine is
designed to make game states compatible with semantic representations,
which bridges the game state space and human language space. And to
realize this, we deploy multi-task learning \cite{DBLP:conf/icml/CollobertW08,DBLP:journals/corr/abs-1811-06031}
in our proposed models.

Data-to-text generation is a popular track in NLG researches. Recent
researches are mainly about generating from structured data to biography
\cite{DBLP:conf/aaai/ShaMLPLCS18}, market comments \cite{DBLP:conf/acl/MurakamiWMGYTM17},
and game reports \cite{DBLP:conf/coling/Li018}. Here we manage to
ground the commentary to the game data (boards and moves). Addressing
content selection \cite{DBLP:conf/emnlp/WisemanSR17} is one of the
top considerations in our designs.

\section{Our Approach}

\begin{figure*}
\begin{centering}
\includegraphics[width=6in]{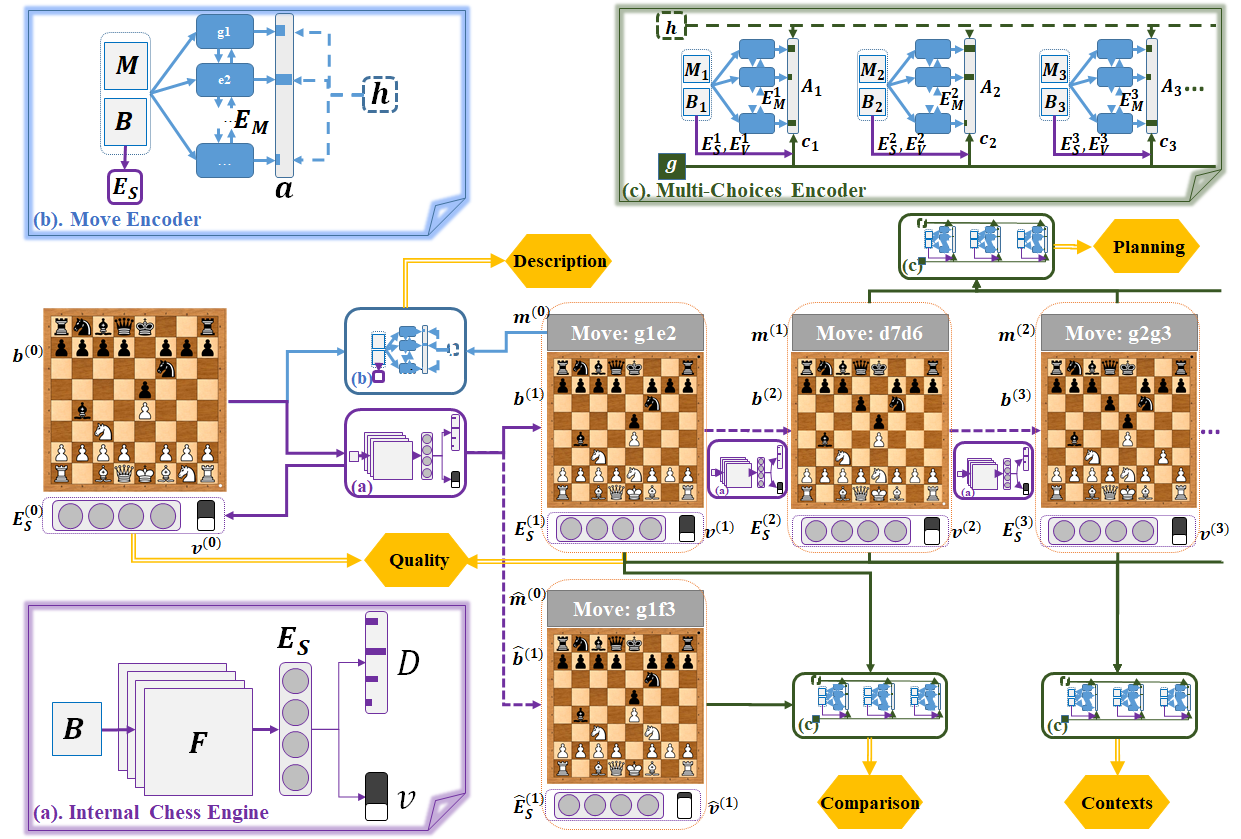} 
\par\end{centering}
\caption{Overview of our chess commentary model.\label{fig:model}}
\end{figure*}

The overview of our approach is shown in Figure \ref{fig:model}.
Apart from the text generation models, there are three crucial modules
in our approach: the internal chess engine, the move encoder, and
the multi-choices encoder. We will first introduce our solution to
all the sub-tasks of chess commentary generation with the modules
as black boxes. And then we describe them in details.

\subsection{Our Solutions}

In Figure \ref{fig:model}, an example is presented with model structures
to demonstrate the way our models solving all the sub-tasks. The process
is impelled by the internal chess engine. Given the current board
$b^{(0)}$ and move $m^{(0)}$, the engine emulates the game and provides
the current and next board states together with wining rates of the
players. Besides, the engine also predicts for another optional move
$\hat{m}^{(0)}$ from $b^{(0)}$ to make comparisons to $m^{(0)}$.
And then a series of long-term moves ($m^{(1)},m^{(2)},...$) and
boards ($b^{(2)},b^{(3)},...$) are further predicted by the engine
in a self-play manner \cite{DBLP:journals/corr/abs-1712-01815,silver2017mastering}
for deep analysis. With the semantics provided by the engine, generation
models are able to predict with abundant and informative contexts.
We will first detail the different semantic contexts with respect
to models for 5 different subtasks. And then we summarize the common
decoding process for all the models.

\textbf{Description Model:} Descriptions about the current move intuitively
depend on the move itself. However, playing the same move could have
different motivations under different contexts. For example, \textit{e2e4}
is the classic Queen Pawn Opening in a fresh start. But it can be
forming a pawn defense structure in the middle of the game. Different
from previous works for chess commentary generation \cite{DBLP:conf/acl/HovyNBGJ18,DBLP:conf/cig/KamekoMT15},
we find all kinds of latent relationships in the current board vital
for current move analysis. Therefore, our description model takes
the representation of both $b^{(0)}$ and $m^{(0)}$ from the move
encoder $f_{ME}$ as semantic contexts to produce description comment
$Y_{Desc}$. The description model is formulated as Eq.\ref{eq:desc}.
\begin{equation}
f_{Description}(f_{ME}(b^{(0)},m^{(0)}))\rightarrow Y_{Desc}\label{eq:desc}
\end{equation}

\textbf{Quality Model: }Jhamtani et al. \shortcite{DBLP:conf/acl/HovyNBGJ18}
find the wining rate features benefit the generation models on \textit{Quality}
category. Inspired by this, we concatenate the current board state
$E_{S}^{(0)}$, the next board state $E_{S}^{(1)}$, and the wining
rate difference $v^{(1)}-v^{(0)}$ as semantic contexts for the decoder.
And to model the value of wining rate difference, we introduce a weight
matrix $W_{diff}$ to map the board state-value pair $[E_{S}^{(0)};E_{S}^{(1)};v^{(1)}-v^{(0)}]$
to the same semantic space of the other contexts by Eq.\ref{eq:diff}.
Our quality model is formulated as Eq.\ref{eq:quality}, where $Y_{Qual}$
is the target comment about quality. 
\begin{equation}
E_{D}=W_{diff}[E_{S}^{(0)};E_{S}^{(1)};v^{(1)}-v^{(0)}]\label{eq:diff}
\end{equation}
\begin{equation}
f_{Quality}(E_{S}^{(0)},E_{S}^{(1)},E_{D})\rightarrow Y_{Qual}\label{eq:quality}
\end{equation}

\textbf{Comparison Model:} Usually, there are more than 10 possible
moves in a given board. But not all of them are worth considering.
Kameko et al. \shortcite{DBLP:conf/cig/KamekoMT15} propose an interesting
phenomenon in chess commentary: when the expert commentators comment
about a bad move, they usually explain why the move is bad by showing
the right move, but not another bad move. Inspired by this, we only
consider the true move $m^{(0)}$ and the potential best move $\hat{m}^{(0)}$
(decided by the internal chess engine) as options for the comparison
model. And the semantic contexts for the options are encoded by the
multi-choices encoder. We define the comparison model as Eq.\ref{eq:comp}
, where $f_{MCE}$ is the multi-choices encoder, $b^{(1)}$ is the
board after executing $m^{(0)}$ on $b^{(0)}$, $\hat{b}^{(1)}$ is
the board after executing $\hat{m}^{(0)}$ on $b^{(0)}$, and $Y_{Comp}$
is the target comment about comparison. 
\begin{multline}
f_{Comparison}(f_{MCE}((b^{(1)},m^{(0)}),(\hat{b}^{(1)},\hat{m}^{(0)})))\\
\rightarrow Y_{Comp}\label{eq:comp}
\end{multline}

\textbf{Planning Model:} We can always find such scenes where commentators
try to predict what will happen assuming they are playing the game.
And then they give analysis according to their simulations. Our internal
chess engine is able to simulate and predict the game in a similar
way (self-play). We realize our model for planning by imitating the
human commentators' behavior. Predicted moves and boards are processed
by our multi-choices encoder to tell the potential big moments in
the future. And we use the multi-choices encoder $f_{MCE}$ to produce
the semantic contexts for the decoder. The process to generate planning
comment $Y_{Plan}$ is described in Eq.\ref{eq:plan}. 
\begin{multline}
f_{Planning}(f_{MCE}((b^{(2)},m^{(1)}),(b^{(3)},m^{(2)}),\\
(b^{(4)},m^{(3)}),...))\rightarrow Y_{Plan}\label{eq:plan}
\end{multline}

\textbf{Contexts Model: }To analyze the situation of the whole game,
the model should know about not only the current, but also the future.
And similar to the planning model, contexts model takes a series of
long-term moves and boards produced by self-play predictions as inputs.
In this way, the model comments the game in a god-like perspective.
And the semantic contexts is also processed by the multi-choices encoder
for generating contexts comment $Y_{Cont}$ as Eq.\ref{eq:context}.
\begin{multline}
f_{Contexts}(f_{MCE}((b^{(1)},m^{(0)}),(b^{(2)},m^{(1)}),\\
(b^{(3)},m^{(2)}),(b^{(4)},m^{(3)}),...))\rightarrow Y_{Cont}\label{eq:context}
\end{multline}

Each of the above models has a decoder (the hexagon blocks in Figure
\ref{fig:model}) for text generation and we use LSTM decoders \cite{sundermeyer2012lstm}.
And we use cross entropy loss function for training. The function
is formalized as Eq.\ref{eq:loss_gen}, where $Y$ is the gold standard
outputs. 
\begin{equation}
Loss_{Gen}=-logp(Y|b^{(0)};m^{(0)})\label{eq:loss_gen}
\end{equation}
We denote $E\in{\rm I\!R}^{n\times d}$ as a bunch of raw context
vectors, where $n$ is the number of such context vectors and $d$
is the dimension of the vectors. Although the semantic contexts $E$
for different generation models are different as described before,
we regard all of the board states, wining rates, and move representations
as general semantic contexts. And we use attention mechanism \cite{DBLP:journals/corr/BahdanauCB14,DBLP:conf/emnlp/LuongPM15}
to gather information from the contexts. For example, assuming that
we have a hidden vector $h$ drawing from LSTM units, to decode with
the semantic contexts, we use the score function $f$ of Luong attention
\cite{DBLP:conf/emnlp/LuongPM15} as 
\begin{equation}
f(X,y)=XWy,\label{eq:luong}
\end{equation}
to calculate the attention weights $a$ for vectors in $E$, where
$W$ is a transformation function for the attentional context vectors.
The scores are further normalized by a softmax function to $a$ by
\begin{equation}
a=\boldsymbol{softmax}(f(E,h)).\label{eq:att_2}
\end{equation}
We compute weighted sum of $E$ with $a$ to produce the attentional
context vector $z$ for word decoding 
\begin{equation}
z=E^{\top}a.\label{eq:att_3}
\end{equation}

\subsection{The Internal Chess Engine}

The internal chess engine is in charge of the mapping from board $B$
to semantic representation $E_{S}$, predicting possibility distribution
$D$ on valid moves, and evaluating the wining rate $v$ for the players.
In previous works \cite{DBLP:conf/acl/HovyNBGJ18,DBLP:conf/cig/KamekoMT15},
researchers use discrete information (threats, game evaluation scores,
etc.) analyzed by external chess engine to build semantic representations.
It limits the capability of the representations by simply mapping
the independent features. Our internal chess engine is able to mine
deeper relations and semantics with the raw board as input. And it
can also make predictions in a continuous semantic space, increasing
the capability and robustness for generation.

Following advanced researches in neural chess engines \cite{DBLP:journals/corr/abs-1711-09667,DBLP:journals/corr/abs-1712-01815},
we split the input raw board into 20 feature planes $F$ for the sake
of machine understanding. There are 12 planes for pieces' (pawn, rook,
knight, bishop, queen, king) positions of each player, 4 planes for
white's repetitions, black's repetitions, total moves, and moves with
no progress, and 4 planes for 2 castling choices of each player. The
feature planes $F$ are encoded by several CNN layers to produce sufficient
information for semantic representation $E_{S}$. Like previous researches
on chess engines, $E_{S}$ is used to predict the move possibility
distribution $D$ and the wining rate $v$ by fully connected layers.
But different from those pure engines, we share the board state $E_{S}$
with generation models in a multi-task manner \cite{DBLP:conf/icml/CollobertW08}.
The engine is designed not only for playing, but also for expressing.
Our generation models use $E_{S}$ as part of the inputs to get better
understanding of the game states.

Given the tuple of game replays $(B,M,v')$ where $M$ is the corresponding
move and $v'$ is the ground truth wining rate, we optimize the engine's
policy, value function at the same time as Eq.\ref{eq:loss_chess_1}
shows. When the engine grows stronger, we let the engine produce data
by itself in a self-play manner \cite{DBLP:journals/corr/abs-1712-01815}.
Besides, the engine jointly optimizes $Loss_{Gen}$ when training
generative models. 
\begin{equation}
Loss_{Eng}=-logp(M|B)+(v-v')^{2}\label{eq:loss_chess_1}
\end{equation}

\subsection{The Move Encoder}

Apart from understanding the board $B$, commentators also need to
know the semantics of the move $M$. Besides using the chess engine
to produce board representations $E_{S}$, the move encoders also
prepare for move embeddings $E_{M}$ as attention contexts for the
text decoders. We set the features of the move (starting cell, the
move ending cell, the piece at the starting cell, the piece at the
ending cell, the promotion state, and the checking state) as a sequential
input to a bi-directional RNN \cite{schuster1997bidirectional}. When
a decoder requests attention contexts for hidden state $h$, the encoder
offers $E=[E_{M};E_{S}]$ to build attentional context vector following
Eq.\ref{eq:att_2} and Eq.\ref{eq:att_3}.

\subsection{The Multi-Choices Encoder}

For \textit{Comparison}, \textit{Planning}, and \textit{Contexts},
there are multiple moves derived from variations and predictions.
The model needs to find the bright spots to describe. To encode these
moves and offer precise information for the generation models, we
propose a multi-choices encoder. Human commentators usually choose
different aspects to comment according to their experiences. We use
a global vector $g$ to store our models' experiences and choose important
moves to comment. Note that $g$ is to be learned. In module (c) of
Figure \ref{fig:model}, we denote $E_{M}^{i}$ as the output vectors
of the $i$-th move encoder, $E_{S}^{i}$ as the board state of the
$i$-th board, and $E_{V}^{i}$ as the embedding of wining rate $v^{i}$
of the $i$-th board. To model the wining rate value, we introduce
a mapping matrix $M_{val}$ and process the state-value pair to the
value embedding as 
\begin{equation}
E_{V}^{i}=W_{val}[E_{S}^{i},v^{i}].\label{eq:val}
\end{equation}
Then we calculate the soft weights of choices $c=\{c_{1},c_{2},...\}$
with respect to the board states $S=\{E_{S}^{1},E_{S}^{2},...\}$
by Eq.\ref{eq:choice}. For hidden state vector $h$ from decoder,
attention weight matrix $A=\{A_{1},A_{2},...\}$ are scaled by $c$
via Eq.\ref{eq:ATT}. And we finally get attentional context vector
$z$ according to $A$ by Eq.\ref{eq:ZZZ}. This approach enables
generation models to generate comments with attention to intriguing
board states. And the attention weights can be more accurate when
$g$ accumulates abundant experiences in training.

\begin{equation}
c=\boldsymbol{softmax}(gS)\label{eq:choice}
\end{equation}
\begin{equation}
A_{i}=c_{i}*\boldsymbol{softmax}(f([E_{M}^{i};E_{S}^{i};E_{V}^{i}],h))\label{eq:ATT}
\end{equation}
\begin{equation}
z=\sum_{i}([E_{M}^{i};E_{S}^{i};E_{V}^{i}])^{\top}A_{i}\label{eq:ZZZ}
\end{equation}

\section{Experiments}

\subsection{Dataset}

We conduct our experiments on recently proposed Chess Commentary dataset\footnote{https://github.com/harsh19/ChessCommentaryGeneration/\label{gac}}
\cite{DBLP:conf/acl/HovyNBGJ18}. In this dataset, Jhamtani et al.
\shortcite{DBLP:conf/acl/HovyNBGJ18} collect and process 11,578
annotated chess games from a large social forum GAMEKNOT\footnote{https://gameknot.com}.
There are 298K aligned data pairs of game moves and commentaries.
The dataset is split into training set, validation set and test set
as a 7:1:2 ratio with respect to the games. As the GAMEKNOT is a free-speech
forum, the comments can be very freewheeling in grammar and morphology.
The informal language style and unpredictable expression tendency
make a big challenge for data-driven neural generation models. To
narrow down the expression tendency, Jhamtani et al. \shortcite{DBLP:conf/acl/HovyNBGJ18}
classify the dataset into 6 categories: \textit{Description}, \textit{Quality},
\textit{Comparison}, \textit{Planning}, \textit{Contexts}, and \textit{General}.
The \textit{General} category is usually about the player and tournament
information, which needs external knowledge irrelevant to game analysis.
We do not conduct experiments on the last category.

And for the training of chess engine, we collect all of the standard
chess game records in the past 10 years from FICS Games Database.
And we remove the games where any player's rating below 2,000. There
are 36M training data (for single move step) after cleaning.

\subsection{Experiment Settings and Baselines}

We train our neural chess engine using mixed data consisting of supervised
FICS data and unsupervised self-play data. The number of self-play
games are set to 0 initially. And it will be increased by 1 when the
trained model beats the previous best version (with a wining rate
larger than 0.55 in 20 games). During 400 iterations of training,
we pick one strong engine and one weak engine for further experiments.
The stronger engine loses 1 game and draws 55 games to the weak engine
in 100 games. As mentioned in Section 3.2, when training generation
models, we use the pre-trained chess engine and fine-tune it with
the generation models.

Here we introduce our models and baselines in the experiments. We
call our models the Skilled Chess Commentator (SCC) as they have the
skills of playing chess. 
\begin{itemize}
\item \textbf{SCC-weak}: The generation models are integrated with the weak
engine mentioned above, and they are trained independently with respect
to the 5 categories in Chess Commentary dataset. 
\item \textbf{SCC-strong}: The model is similar to SCC-weak, but integrated
with the strong engine. 
\item \textbf{SCC-mult}: This is a multi-task learning model where generation
models for different categories share the strong chess engine, move
encoder, the multi-choices encoder and the value mapping matrix $W_{val}$. 
\item \textbf{GAC}: The state-of-the-art method proposed by Jhamtani et
al. \shortcite{DBLP:conf/acl/HovyNBGJ18}. Their models incorporate
the domain knowledge provided by external chess engines. Their models
only work for first 3 categories: \textit{Description}, \textit{Quality},
and \textit{Comparison}. We will compare our results with \textbf{GAC}
on these categories. 
\item \textbf{KWG}: Another state-of-the-art method for game commentary
generation \cite{DBLP:conf/cig/KamekoMT15}. It is a pipeline method
based on keyword generation. We compare the results on all data categories. 
\item \textbf{Temp}: This is a template-based baseline methods. Together
with the dataset, Jhamtani et al. \shortcite{DBLP:conf/acl/HovyNBGJ18}
provide templates for the first two categories. Inspired by \cite{DBLP:conf/cg/SadikovMGKB06},
we extend the templates to fit for all the 5 categories. 
\item \textbf{Re}: This is a retrieval-based baseline method. For each input
in the test set, we find the most matched datum in the training set
by numbers of matched input board and move features. 
\end{itemize}
\begin{table*}[t]
\begin{centering}
\begin{tabular}{c||c|c||c|c||c|c|c}
\hline 
\textbf{BLEU-4 (\%)}  & \textbf{Temp}  & \textbf{Re}  & \textbf{KWG}  & \textbf{GAC}  & \textbf{SCC-weak}  & \textbf{SCC-strong}  & \textbf{SCC-mult}\tabularnewline
\hline 
\textbf{Description}  & 0.82  & 1.24  & 1.22  & \textbf{1.42}  & 1.23  & 1.31  & 1.34\tabularnewline
\textbf{Quality}  & 13.71  & 4.91  & 13.62  & 16.90  & 16.83  & 18.87  & \textbf{20.06}\tabularnewline
\textbf{Comparison}  & 0.11  & 1.03  & 1.07  & 1.37  & 2.33  & \textbf{3.05}  & 2.53\tabularnewline
\textbf{Planning}  & 0.05  & 0.57  & 0.84  & N/A  & \textbf{1.07}  & 0.99  & 0.90\tabularnewline
\textbf{Contexts}  & 1.94  & 2.70  & 4.39  & N/A  & 4.04  & \textbf{6.21}  & 4.09\tabularnewline
\hline 
\hline 
\textbf{BLEU-2 (\%)}  & \textbf{Temp}  & \textbf{Re}  & \textbf{KWG}  & \textbf{GAC}  & \textbf{SCC-weak}  & \textbf{SCC-strong}  & \textbf{SCC-mult}\tabularnewline
\hline 
\textbf{Description}  & 24.42  & 22.11  & 18.69  & 19.46  & 23.29  & \textbf{25.98}  & 25.87\tabularnewline
\textbf{Quality}  & 46.29  & 39.14  & 55.13  & 47.80  & 58.53  & 61.13  & \textbf{61.62}\tabularnewline
\textbf{Comparison}  & 7.33  & 22.58  & 20.06  & 24.89  & 24.85  & \textbf{27.48}  & 23.47\tabularnewline
\textbf{Planning}  & 3.38  & 20.34  & 22.02  & N/A  & 22.28  & \textbf{25.82}  & 24.32\tabularnewline
\textbf{Contexts}  & 26.03  & 30.12  & 31.58  & N/A  & 37.32  & \textbf{41.59}  & 38.59\tabularnewline
\hline 
\hline 
\textbf{METEOR (\%)}  & \textbf{Temp}  & \textbf{Re}  & \textbf{KWG}  & \textbf{GAC}  & \textbf{SCC-weak}  & \textbf{SCC-strong}  & \textbf{SCC-mult}\tabularnewline
\hline 
\textbf{Description}  & 6.26  & 5.27  & 6.07  & 6.19  & 6.03  & 6.83  & \textbf{7.10}\tabularnewline
\textbf{Quality}  & 22.95  & 17.01  & 22.86  & 24.20  & 24.89  & \textbf{25.57}  & 25.37\tabularnewline
\textbf{Comparison}  & 4.27  & 8.00  & 7.70  & 8.54  & 8.25  & \textbf{9.44}  & 9.13\tabularnewline
\textbf{Planning}  & 3.05  & 6.00  & 6.76  & N/A  & 6.18  & 7.14  & \textbf{7.30}\tabularnewline
\textbf{Contexts}  & 9.46  & 8.90  & 10.31  & N/A  & 11.07  & \textbf{11.76}  & 11.09\tabularnewline
\hline 
\end{tabular}
\par\end{centering}
\caption{Automatic evaluation results.}
\label{auto_eval} 
\end{table*}

\subsection{Evaluation Metrics}

We develop both automatic evaluations and human evaluations to compare
the models.

For automatic evaluations, we use BLEU \cite{papineni2002bleu} and
METEOR \cite{denkowski:lavie:meteor-wmt:2014} to evaluate the generated
comments with ground-truth outputs. BLEU evaluates the modified precision
between the predicted texts and gold-standard references on corpus
level. Evaluating with 4-grams (BLEU-4 \footnote{https://github.com/moses-smt/mosesdecoder/blob/ master/scripts/generic/multi-bleu.perl})
is the most popular way in NLG researches. However, for tasks like
dialogue system \cite{DBLP:conf/naacl/LiGBGD16}, story telling generation
\cite{DBLP:journals/corr/JainAMSLS17}, and chess commentary \cite{DBLP:conf/acl/HovyNBGJ18},
the outputs can be rather short and free expressions. Under such circumstances,
brevity penalty for 4-grams can be too strict and makes the results
unbalanced. We use BLEU-2 \footnote{https://github.com/harsh19/ChessCommentaryGeneration/ blob/master/Code/methods/category\_aware/BLEU2.perl}
to show more steady results with BLEU evaluation algorithm. We also
use METEOR as a metric, whose results are more closed to a normal
distribution \cite{dobre2015comparison}.

We also conduct human evaluation to make more convincing comparisons.
We recruit 10 workers on Amazon Mechanical Turk\footnote{https://www.mturk.com}
to evaluate 150 groups of samples (30 from each category). Each sample
is assigned to exactly 2 workers. The workers rate 8 shuffled texts
(for \textbf{Ground Truth}, \textbf{Temp}, \textbf{Re}, \textbf{GAC},
\textbf{KWG}, and \textbf{SCC} models) for the following 4 aspect
in a 5-pt Likert scale\footnote{https://en.wikipedia.org/wiki/Likert scale}. 
\begin{itemize}
\item \textbf{Fluency}: Whether the comment is fluent and grammatical. 
\item \textbf{Accuracy}: Whether the comment correctly describes current
board and move. 
\item \textbf{Insights}: Whether the comment makes appropriate predictions
and thorough analysis. 
\item \textbf{Overall}: The annotators' overall impression about comments. 
\end{itemize}
\begin{figure*}
\centering{}\includegraphics[width=6in]{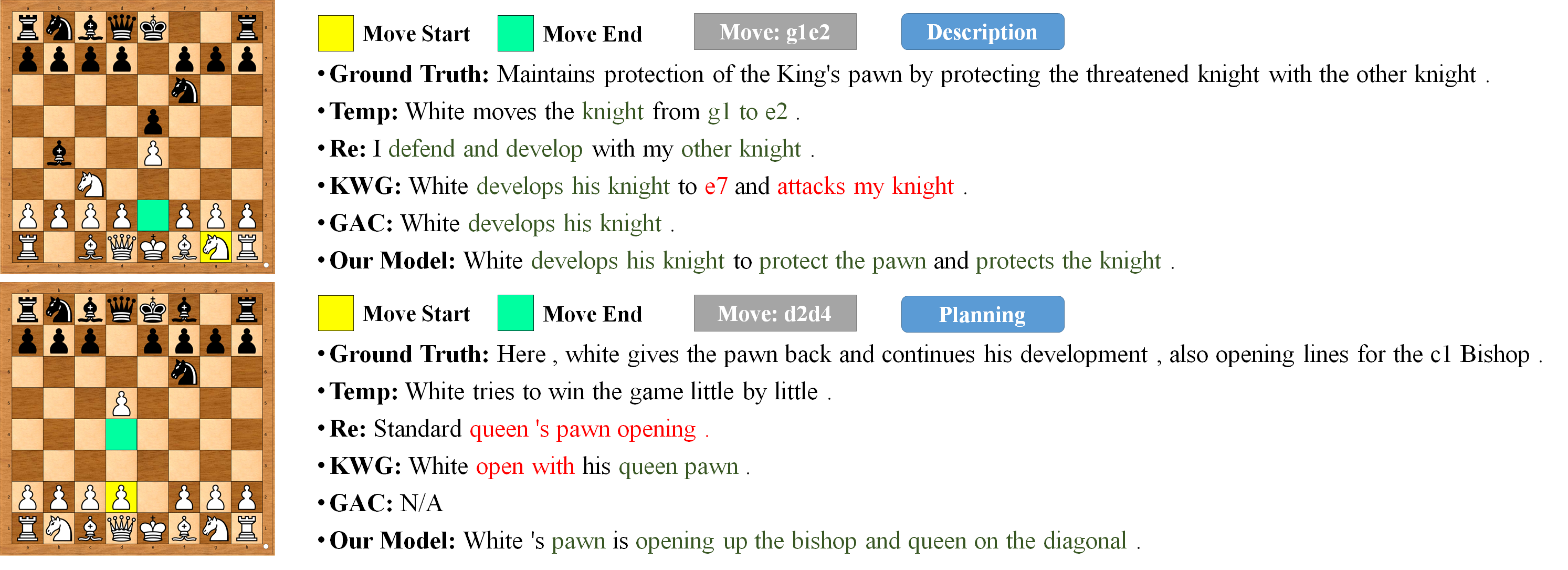}\caption{Samples for case study.\label{fig:Samples-for-case}}
\end{figure*}

\subsection{Results and Analysis}

We present the automatic evaluation results in Table \ref{auto_eval}.
Our \textbf{SCC} models outperform all of the baselines and previous
state-of-the-art models. \textbf{Temp} is limited by the variety of
templates. It is competitive with the neural models on \textit{Description}
and \textit{Quality} due to limited expressions in these tasks. But
when coming to \textit{Comparison}, \textit{Planning} and \textit{Contexts},
\textbf{Temp} shows really bad performances. \textbf{Re} keeps flexibility
by copying the sentences from training set. But it does not perform
well, either. The ability of \textbf{Re} is limited by the sparse
searching space, where there are 90,743 data in the training set,
but $10^{43}$ possible boards\footnote{https://en.wikipedia.org/wiki/Shannon\_number}
for chess game. \textbf{KWG} and \textbf{GAC} provide competitive
results. With the help of external information from powerful chess
engines, \textbf{GAC} shows good performances on \textit{Quality}
and \textit{Comparison}. Although our internal chess engine is no
match for the external engines that \textbf{GAC} uses at playing chess,
it turns out that our models with directly internal information can
better bridge the semantic spaces of chess game and comment language.
As for the comparisons within our models, \textbf{SCC-strong} turns
to be better than \textbf{SCC-weak}, which supports our assumption
that better skills enable more precise predictions, resulting in better
comments. Training with multi-task learning seems to hurt the overall
performances a little. But \textbf{SCC-mult} still has the state-of-the-art
performances. And more important, it can react to all sub-tasks as
a whole.

The human annotators are required to be good at playing chess. That
is to say, they are the true audiences of the commentator researches
and applications. By introducing human evaluations, we further reveal
the performances in the perspective of the audiences. We show the
average scores and significance test results in Table \ref{human_eval}.
We further demonstrate the efficacy of our models with significantly
better overall performances than the retrieval-based model and previous
state-of-the-art ones. It is worth noting that the evaluations about
Accuracy and Insights show that our models can produce more precise
and thorough analysis owing to the internal chess engine. \textbf{SCC-mult}
and \textbf{SCC-strong} perform better than \textbf{SCC-weak} in Accuracy
and Overall scores. It also supports the points that the our commentary
model can be improved with better internal engine. 
\begin{table}[t]
\begin{centering}
\resizebox{3in}{0.85in}{%
\begin{tabular}{c|c|c|c|c}
\hline 
\textbf{Models}  & \textbf{Fluency}  & \textbf{Accuracy}  & \textbf{Insights}  & \textbf{Overall}\tabularnewline
\hline 
\hline 
\textbf{Ground Truth}  & \textbf{4.02}  & \textbf{3.88}  & \textbf{3.58}  & \textbf{3.84}\tabularnewline
\hline 
\hline 
\textbf{Temp}  & \textbf{4.05}  & \textbf{4.03}  & \uline{3.02}  & 3.56\tabularnewline
\textbf{Re}  & 3.71  & \uline{3.00}  & \uline{2.80}  & \uline{2.85}\tabularnewline
\hline 
\textbf{KWG}  & \uline{3.51}  & \uline{3.24}  & \uline{2.93}  & \uline{3.00}\tabularnewline
\hline 
\textbf{SCC-weak}  & 3.63  & \uline{3.62}  & 3.32  & \uline{3.30}\tabularnewline
\textbf{SCC-strong}  & 3.81  & 3.74  & 3.49  & 3.49\tabularnewline
\textbf{SCC-mult}  & 3.82  & 3.91  & \textbf{3.51}  & \textbf{3.61}\tabularnewline
\hline 
\hline 
\textbf{GAC{*}}  & 3.68  & \uline{3.32}  & \uline{2.99}  & \uline{3.14}\tabularnewline
\hline 
\textbf{SCC-mult{*}}  & 3.83  & 3.99  & 3.46  & 3.52\tabularnewline
\hline 
\end{tabular}} 
\par\end{centering}
\caption{Human evaluation results. Models marked with \textbf{{*}} are evaluated
only for the \textit{Description}, \textit{Quality}, and \textit{Comparison}
categories. The underlined results are significantly worse than those
of\textbf{ SCC-mult({*}) }in a two-tail T-test (p\textless 0.01).}

\label{human_eval} 
\end{table}

\subsection{Case Study}

To have a better view of comparisons among model outputs, we present
and analyze some samples in Figure \ref{fig:Samples-for-case}. In
these samples, our model refers to \textbf{SCC-mult}.

For the first example, black can exchange white's \textit{e3} knight
and \textit{e4} pawn with the \textit{b4} bishop if white takes no
action. But white chooses to protect the \textit{e3} knight with the
\textit{g1} knight. All the models generate comments about \textit{Description}.
\textbf{Temp} directly describes the move without explanation. \textbf{Re}
finds similar situation in the training set and explains the move
as defense and developing. \textbf{KWG} is right about developing,
but wrong about the position of the knight and the threats. \textbf{GAC}
produces safe comment about the developing. And our model has a better
understanding about the boards. It annotates the move correctly and
even gives the reason why white plays this move.

For the second example, the game is at the 3rd turn. White gives up
the pawn on \textit{d5} and chooses to push the queen's pawn. \textbf{Re}
and \textbf{KWG} both make a mistake and recognize the move \textit{d2d4}
as Queen Pawn Opening. \textbf{Temp} thinks white is going to win
because white have the advantage of one more pawn. However, \textbf{Temp}
cannot predict that white will lose the advantage in the next move.
Our model is able to predict the future moves via self-play. And it
draws the conclusion that pushing the queen's pawn can open up the
ways for the queen and bishop for future planning.

\section{Conclusion and Future Work}

In this work we propose a new approach for automated chess commentary
generation. We come up with the idea that models capable of playing
chess will generate good comments, and models with better playing
strength will perform better in generation. By introducing a compatible
chess engine to comment generation models, we get models that can
mine deeper information and ground more insightful comments to the
input boards and moves. Comprehensive experiments demonstrate the
effectiveness of our models.

Our experiment results show the direction to further developing the
state-of-the-art chess engine to improve generation models. Another
interesting direction is to extend our models to multi-move commentary
generation tasks. And unsupervised approaches to leverage massive
chess comments in social media is also worth exploring.

\section*{Acknowledgments}

This work was supported by National Natural Science Foundation of
China (61772036) and Key Laboratory of Science, Technology and Standard
in Press Industry (Key Laboratory of Intelligent Press Media Technology).
We thank the anonymous reviewers for their helpful comments. Xiaojun
Wan is the corresponding author.

\bibliography{acl2019}
 \bibliographystyle{acl_natbib} 
\end{document}